\def\year{2018}\relax
\documentclass[letterpaper]{article} 
\usepackage{aaai18}  
\usepackage{times}  
\usepackage{helvet}  
\usepackage{courier}  
\usepackage{url}  
\usepackage{graphicx}  
\newcommand{\figref}[1]{Figure~\ref{fig:#1}}

\begin{document}
%
\title{Personalized Human Activity Recognition Using Convolutional Neural Networks}
\author{Seyed Ali Rokni, Marjan Nourollahi, \and Hassan Ghasemzadeh\\
Washington State University\\
School of Electrical Engineering and Computer Science\\
Pullman, Washington 99164--2752\\
\{alirokni,mnourol,hassan\}@eecs.wsu.edu
}

\maketitle
\begin{abstract}
A major barrier to the personalized Human Activity Recognition using wearable sensors is that the performance of the recognition model drops significantly upon adoption of the system by new users or changes in physical/ behavioral status of users. Therefore, the model needs to be retrained by collecting new labeled data in the new context. In this study, we develop a transfer learning framework using convolutional neural networks to build a personalized activity recognition model with minimal user supervision.  
\end{abstract}

\section{Introduction}


Inertial wearable sensors have been vastly utilized for Human Activity Recognition (HAR). A major challenge with the trained HAR models is that the performance of the classifier is highly sensitive to the context of the sensor and  engineered features \cite{rokni2017synchronous}. Upon any changes in the task or distribution of the data  (e.g., a new user utilizing the system, or changes in activities of interest), we will need to obtain additional inputs from a human expert by redoing the costly process of collecting labeled data and handcrafting features. 
This problem becomes more challenging considering that wearables are deployed in highly dynamic and uncontrolled environments, mainly due to their direct and continuous exposure to end-users and their living environments. 


To avoid handcrafting features, the growing trend of representation learning from raw sensor data with Convolution Neural Networks (ConvNets) has demonstrated a great performance in activity recognition in different domains \cite{zeng2014convolutional,yang2015deep,ronao2016human}.  
 

To expand the pattern recognition capabilities from a single setting algorithm with a predefined configuration to a dynamic setting, successful knowledge transfer is needed to improve the learning performance by avoiding expensive data collection, labeling and training efforts. Authors of \cite{bengio2011deep} showed that deep learners are more powerful in utilizing data points that are not from the same distribution as the training distribution of a shallow learner. Particularly, \cite{yosinski2014transferable} demonstrated that features learned in the first layers are not specific to a particular task and could be useful for other related tasks. 

In this study, we develop a neural network architecture which enables us to build a personalized HAR model with minimal human supervision.

\section{Representation Learning for Sensory Data}\label{sec:replear}

A 3D accelerometer sensor captures a sample of body acceleration is the form of  \begin{equation}
v_t = [v_t^x ; v_t^y ; v_t^z]
\end{equation}  
where $v_t^x$, $v_t^y$ and $v_t^z$ denote x, y and z-acceleration, respectively.


Because the sensor captures human accelerations continuously while the subject performs different activities in free-living situations, `start' and `end' of activities are unknown {\it a priori}. 
A typical segmentation with a window of size $w$ on 3-axis accelerometer data forms 3 channels of input data, $C_t = [C_x^t C_y^tC_z^t]$, where
\begin{equation}
C_x^t = [v_t^x, \dots , v_{t+w-1}^x] 
\end{equation}
\begin{equation}
C_y^t = [v_t^y, \dots , v_{t+w-1}^y] 
\end{equation}
\begin{equation}
C_z^t = [v_t^z, \dots , v_{t+w-1}^z] 
\end{equation}

Assuming $M$ activities of interest $\mathcal{A}$=\{$a_1$, $a_2$, $\dots$, $a_M$\}, the activity recognition task assigns label $a_j \in \mathcal{A}$ to an observed segment $C_t$. 



The input layer of our neural network structure consists of 3 channels of the smoothed signal segment. First, these segments are passed through discretization layer. The discretization helps to reduce sensitivity of the model to small changes and therefore makes the model more robust in transferring into other domains. 
Typically, a small portion of the signal range is used by an activity segment and this range varies from one activity to another. Therefore, we feed the discretized sequence to an embedding layer to generate a compact representation of the input data. In addition to dimensionality reduction, this embedding layer could be useful  to reduce the effect of different instrumental calibration. Therefore, similar activities captured by different accelerometer sensors or performed by different users could have close representation in this space. 
The next layer is a stack of 1D convolutional layers. Each convolutional layer captures local dependencies and scales invariant characteristics of the input. 
The sparse connectivity and parameter sharing features of ConvNets not only help in extracting useful local features on different body locations, but also reduce the computational and storage complexity of the model which is an essential consideration for algorithms that run on embedded sensory devices. A ReLU activation function is applied on the linear output each of convolutional layer followed by a max pooling function that replaces the output of the unit with the maximum output of nearby units. Because the pooling summarizes the outputs over an entire neighborhood, the pooling layer makes the representation smaller, more manageable, and invariant to local translations. Next, we apply Dropout method to prevent overfitting with computational efficient regularization. Then, on top of output of the last convolutional layer, we add a densely connected layer referred to as {\it Classification Layer} to aggregate all outputs and construct a scoring function. 

%
%
%

\section{Personalized Model using Transfer Learning} 
Having a trained network for a group of users as the source domain, we devise a personalized model by reusing the lower layers of the network and retrained the upper layer with few number of instances in the target domain. Particularly, when a new user utilizes the model, we freeze all layers of the trained network except for the {\it classification layer}. Acquiring a few number of labels for the new user, in multiple epochs, we adjust the weights of the top {\it classification layer} to be more specific to the activity pattern of the current user (i.e., target domain). We call this transfer learning method as Transfer Convolutional (TrC).
  
To evaluate our method, we apply the proposed approach on 2 publicly available datasets, including Sport and Daily Activity (SDA) \cite{altun2010comparative} and WISDM \cite{kwapisz2011activity}, which contain data from multiple users and with multiple activities. For the SDA dataset, where subjects have worn 5 inertial sensors on different body locations, we combine 3D accelerometer channels of each sensor and form an input with 15 channels. 

\begin{figure}[th]
\hspace{-2mm}
\includegraphics[width=0.40\textwidth]{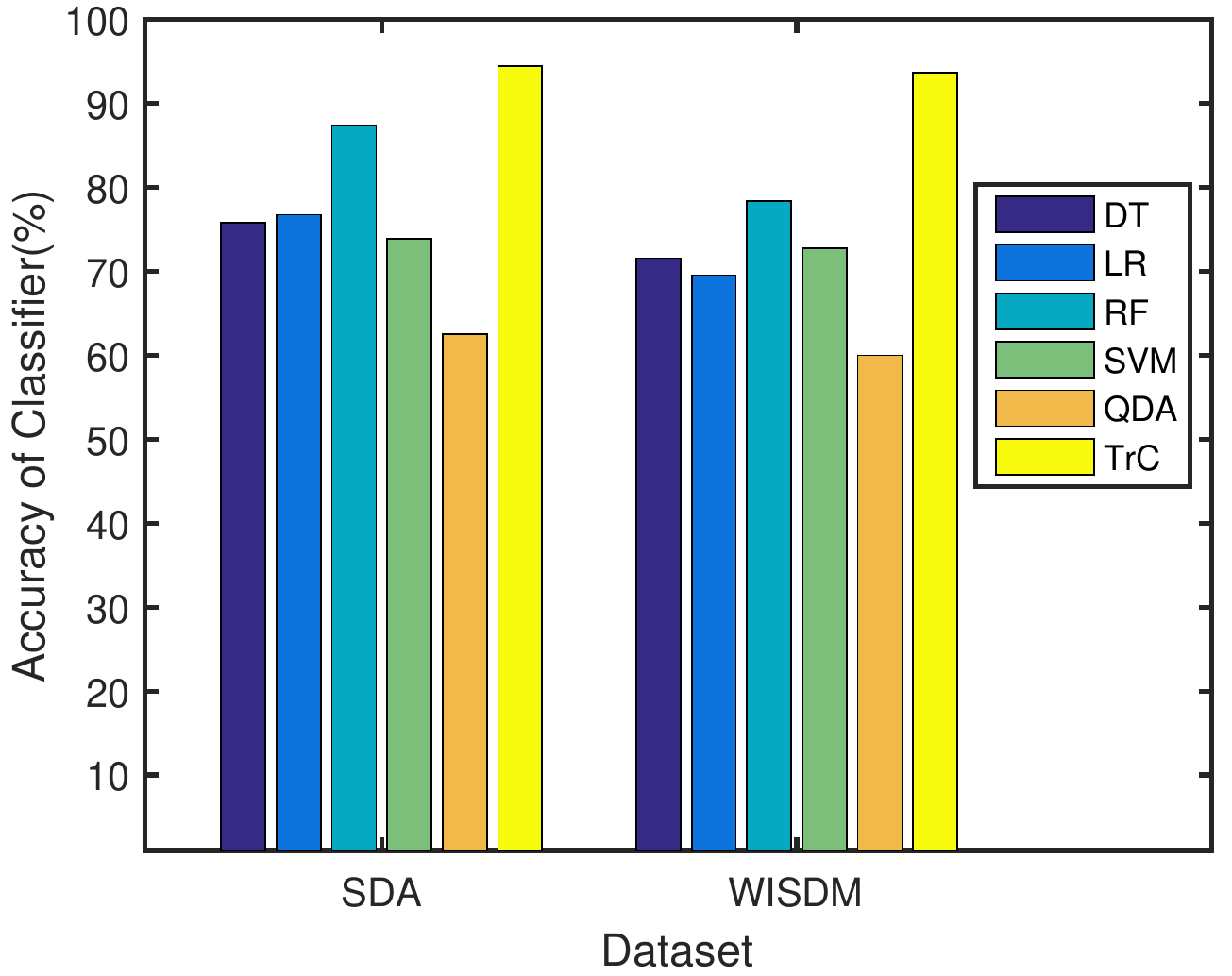}\caption{Performance of different classifiers}
\label{fig:result}
\end{figure}

The validation process is leave-one-subject-out where a user is selected for test and the model is trained on data collected from the remaining subjects. Then, using data associated with the test subject, we build a set of transfer instances by randomly acquiring 3 labeled instances for each activity and retraining the classification layer. Furthermore, we train 5 shallow classifiers Decision Tree (DT), Logistic Regression (LR), Random Forest (RF), SVM ,and Quadratic Discriminant Analysis (QDA) on the combined dataset of training data from other subjects and the transferred instances. All trained models are evaluated on the remaining instances of test subjects (excluding transfer instances). \figref{result} shows averaged performance of the classifiers over all possible leave-one-subject-out scenarios. In this experiment, the network architecture only stacked two layers of convolution and max pooling. As presented in  \figref{result}, our transfer learning approach significantly improves the accuracy of activity recognition with only few labeled instances. 
\section{Conclusion}
Transfer learning could help to adjust already trained activity recognition model for a new user with minimal human supervision. We showed that using representation learning, we can reuse the general features learned from available training data and construct a personalized model with only few labeled instances.
\bibliography{refs}
\bibliographystyle{aaai}

\end{document}